%% file: eacl2023.tex
\pdfoutput=1

\documentclass[11pt]{article}

\usepackage[]{eacl2023}

\usepackage{times}
\usepackage{latexsym}

\usepackage[T1]{fontenc}

\usepackage[utf8]{inputenc}

\usepackage{microtype}

\usepackage{inconsolata}

\usepackage{array}

\usepackage{graphicx}
\usepackage{multirow}
\usepackage{tabularx}
\usepackage{siunitx}
\usepackage{fullpage}
\usepackage{makecell}
\usepackage{booktabs}
\usepackage{arydshln}
\usepackage{float}
\usepackage{nicematrix}
\usepackage{xspace,mfirstuc,tabulary}
\usepackage{cleveref}
\usepackage{bm}
\usepackage{algpseudocode}
\usepackage{algorithm}
\usepackage{siunitx}
\usepackage{enumitem}
\usepackage{url}
\usepackage{bbding}
\usepackage{pifont}
\usepackage{wasysym}
\usepackage{amssymb}
\usepackage{xcolor}
\usepackage{subcaption}

\newcolumntype{C}{>{\centering\arraybackslash}X}

\newcommand{\ignore}[1]{}

\newcommand{\shuai}[1]{\textcolor{blue}{\textbf{[#1 --\textsc{Shuai}]}}}

\newcommand{\robert}[1]{\textcolor{violet}{\textbf{[#1 --\textsc{Robert}]}}}
\newcommand{\smara}[1]{\textcolor{brown}{\textbf{[#1 --\textsc{Smara}]}}}
\newcommand{\miguel}[1]{\textcolor{yellow}{\textbf{[#1 --\textsc{Miguel}]}}}

\title{Instruction Tuning for Few-Shot Aspect-Based Sentiment Analysis}

\author{Siddharth Varia$^1$\thanks{~~Indicates equal contribution.} ~~~~ Shuai Wang$^1$\footnotemark[1] ~~~~ Kishaloy Halder$^1$\footnotemark[1] ~~~Robert Vacareanu$^2$\thanks{~~Work done during internship at AWS.} \footnotemark[1] \\ \textbf{Miguel Ballesteros$^1$} ~~~~ \textbf{Yassine Benajiba$^1$} ~~~~\textbf{Neha Anna John$^1$} \\ ~~~~\textbf{Rishita Anubhai$^1$}  ~~~~\textbf{Smaranda Muresan}$^{1}$ ~~~~\textbf{Dan Roth$^{1}$}  
 \\  $^1$AWS AI Labs \\ $^2$University of Arizona, Tucson, AZ, USA\\
{\tt \footnotesize{\{siddhvar, wshui, kishaloh, ballemig, benajiy, ranubhai, smaranm, drot\}@amazon.com}} 
\\ {\tt \footnotesize{rvacareanu@arizona.edu}} 
 } 

\begin{document}
\maketitle
\begin{abstract}
\input{abstract}
\end{abstract}

\input{introduction}
\input{methods}
\input{experiments}
\input{conclusion}
\input{limitations}

\bibliographystyle{acl_natbib}
\bibliography{eacl2023}

\appendix
\clearpage
\input{appendix}

\end{document}

%% file: abstract.tex
Aspect-based Sentiment Analysis (ABSA) is a fine-grained sentiment analysis task which 
involves 
four elements from user-generated texts: 
aspect term, aspect category, opinion term, and sentiment polarity. Most computational approaches focus on some of the ABSA sub-tasks such as tuple (aspect term, sentiment polarity) or triplet (aspect term, opinion term, sentiment polarity) extraction using either pipeline or joint modeling approaches. 
Recently, generative approaches have been proposed to extract all four elements as (one or more) quadruplets from text as a single task. 
In this work, we take a step further and propose a unified framework for solving ABSA, and the associated sub-tasks to improve the performance in few-shot scenarios. To this end, we fine-tune a {\tt T5} model with \emph{instructional prompts} in a multi-task learning fashion covering all the sub-tasks, as well as the entire quadruple prediction task.
In experiments with multiple benchmark datasets, \ignore{\miguel{by finetuning a T5... },} 
we show that the proposed multi-task prompting approach   
brings 
performance boost (by absolute $8.29$ F1)\ignore{\shuai{averaged over all k-shot setting and datasets?}}
in the few-shot learning setting.

%% file: introduction.tex
\section{Introduction}
\label{sec:intro}
Aspect-Based Sentiment Analysis (ABSA) 
is a fine-grained sentiment analysis task 
where the goal is to extract the sentiment associated with an entity and all its  aspects \cite{liu2012, pontiki-etal-2014-semeval, pontiki-etal-2015-semeval,pontiki-etal-2016-semeval,schouten2015survey,zhang2018deep,nazir2020issues,zhang2022survey}. 
For example, in the context of Restaurant reviews 
the relevant aspects 
could be \textit{food, ambience, location, service} with \textit{general} used to represent the subject itself (i.e., restaurant). 
ABSA can provide valuable fine-grained information for businesses to  
analyze the aspects they care about. 
Annotated datasets have been released to foster research in this area
\cite{pontiki-etal-2014-semeval, pontiki-etal-2015-semeval,pontiki-etal-2016-semeval}.  

\begin{figure}[]
    \centering
    \includegraphics[height=0.85in]{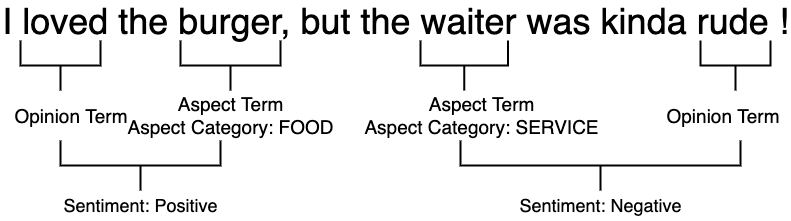}
    \caption{Illustrative orientation of four ABSA elements \textit{i.e.,} Aspect Term, Aspect Category, Opinion Term, and Sentiment. The related tasks often involve predicting either everything together or a subset of them.\ignore{\shuai{Can we integrate the information shown here into Figure 1? }}}
    \label{fig:absa_illustration}
\end{figure}

A full ABSA task aims to extract four elements from a user-generated text: aspect term, aspect category, opinion term and the sentiment polarity (see Figure \ref{fig:absa_illustration} for an example). 
Most existing approaches 
have the focus on extracting some of these elements such as a single element (\textit{e.g.,} aspect term), tuple (\textit{e.g.,} aspect term, sentiment polarity), or triplet (\textit{e.g.,} aspect term, aspect category, sentiment polarity) \cite{li2020conditional,hu2019open,xu2020position}. Recently, \newcite{zhang-etal-2021-aspect-sentiment} tackled the full ABSA task, under the name of Aspect Sentiment Quadruple Prediction (ASQP). Technically, most existing computational approaches 
have used extractive and discriminative models either in a pipeline or in an end-to-end framework \cite{wang2016recursive, yu2018, cai-etal-2021-aspect} to address ABSA. 
Generative approaches have been recently shown to be effective for the full ABSA task and its sub-tasks \cite{zhang-etal-2021-aspect-sentiment,zhang-etal-2021-towards-generative,yan-etal-2021-unified}. Most notably, \newcite{zhang-etal-2021-aspect-sentiment} used a sequence-to-sequence (seq-to-seq) model to address ASQP as a paraphrase generation problem.
One important consideration is that modeling ABSA in a generative fashion allows for cross-task knowledge transfer. \ignore{\shuai{This sentence seems a bit disconnect }}

We go a step further and propose a unified model that can tackle multiple ABSA sub-tasks, including the ASQP task, 
and explore its effectiveness for low data scenarios.
Recent work on large language models relies on the intuition that most natural language processing tasks can be described via natural language instructions and that models trained on these instructions show strong zero-shot performance on several tasks \cite{wei2021finetuned,sanh2022multitask}. 
Based on this success, we propose a unified model based on multi-task prompting with instructional prompts using T5 \cite{JMLR:v21:20-074} to solve the full ABSA task \textit{i.e.,} ASQP \cite{zhang-etal-2021-aspect-sentiment} and several of its associated sub-tasks addressed in the literature: 1) Aspect term Extraction (AE) \cite{jakob2010extracting}; 
2) Aspect term Extraction and Sentiment Classification (AESC)
\cite{yan-etal-2021-unified}; 3) Target Aspect Sentiment Detection (TASD), which aims to extract the aspect term, aspect category, and sentiment polarity \cite{Wan_Yang_Du_Liu_Qi_Pan_2020}; 4) Aspect Sentiment Triplet Extraction (ASTE), which aims to extract the aspect term, opinion term, sentiment polarity \cite{Peng_Xu_Bing_Huang_Lu_Si_2020}
. We conduct an extensive set of experiments with multiple review datasets. 
Experimental results show that our proposed model achieves substantial improvement ($8.29$ F1 on average) against the state-of-the-art in few-shot learning scenario\footnote{Sources available at: \url{https://github.com/amazon-science/instruction-tuning-for-absa}}. 

%% file: methods.tex
\section{Methods}
\label{sec:methods}

The four elements of ABSA form a quadruple as the sentiments are associated with both the aspect, and the opinion terms (\textit{cf} Figure \ref{fig:absa_illustration}). In this work, we hypothesize that it is important to capture the interaction between these components not only at the quadruple level, but also within a subset of these four elements.


\begin{table*}[t]
\centering
\resizebox{1.0\textwidth}{!}{%
\begin{tabular}{ccccclc}
\hline
Task & {\color{red}{\$AT}} & {\color{blue}{\$AC}} & {\color{violet}{\$S}} & {\color{orange}{\$OT}} & \makecell[c]{Input Instruction} & Output \\ \hline
\begin{tabular}[c]{@{}c@{}}Aspect \\Extraction (AE)\end{tabular} & \checkmark & & & & \begin{tabular}[l]{@{}l@{}}Given the text: \$TEXT, what are the aspect terms in it ?\\ What are the aspect terms in the text: \$TEXT ?\end{tabular} & \begin{tabular}[c]{@{}c@{}}Template: {\color{red}{\$AT}}\\Literal: {\color{red}{\textit{burger}}}\end{tabular} \\ \hline
\begin{tabular}[c]{@{}c@{}}Aspect term\\Extraction and\\Sentiment Classification\\(AESC)\end{tabular} & \checkmark & & \checkmark & & \begin{tabular}[l]{@{}l@{}}Given the text: \$TEXT, what are the aspect terms and \\their sentiments ?\\ What are the aspect terms and their sentiments in \\the text: \$TEXT ?\end{tabular} & \begin{tabular}[c]{@{}c@{}} Template: {\color{red}{\$AT}} is {\color{violet}\$S}\\Literal: \textit{{\color{red}burger} is {\color{violet} great}}\end{tabular} \\ \hline
\begin{tabular}[c]{@{}c@{}}Target Aspect \\ Sentiment Detection \\ (TASD)\end{tabular} & \checkmark & \checkmark & \checkmark & & \begin{tabular}[l]{@{}l@{}}Given the text: \$TEXT, what are the aspect terms, \\sentiments and categories ?\\ What are the aspect terms, sentiments and categories \\in the text: \$TEXT ?\end{tabular} & \begin{tabular}[l]{@{}l@{}} Template: {\color{red}\$AT} is {\color{violet}\$S} means \\{\color{blue}\$AC} is {\color{violet}\$S} \\Literal: \textit{{\color{red}burger} is {\color{violet}great} means} \\ \textit{{\color{blue}food} is {\color{violet}great}}\end{tabular} \\ \hline
\begin{tabular}[c]{@{}c@{}}Aspect Sentiment \\Triplet
Extraction\\ (ASTE)\end{tabular} & \checkmark &  & \checkmark & \checkmark & \begin{tabular}[l]{@{}l@{}}Given the text: \$TEXT, what are the aspect terms, \\opinion terms and sentiments ?\\ What are the aspect terms, opinion terms and \\sentiments in the text: \$TEXT ?\end{tabular} & \begin{tabular}[l]{@{}l@{}} Template: {\color{red}\$AT} is {\color{orange}\$OT} means \\it is {\color{violet}\$S} \\ Literal: \textit{{\color{red}burger} is {\color{orange}loved} means} \\ \textit{it is {\color{violet}great}}\end{tabular} \\ \hline
\begin{tabular}[c]{@{}c@{}}Aspect Sentiment \\Quadruple
Prediction\\ (ASQP)\end{tabular} & \checkmark & \checkmark & \checkmark & \checkmark & \begin{tabular}[l]{@{}l@{}}Given the text: \$TEXT, what are the aspect terms, \\opinion terms, sentiments and categories ?\\ What are the aspect terms, opinion terms, sentiments and \\categories in the text: \$TEXT ?\end{tabular} & \begin{tabular}[l]{@{}l@{}}Template: {\color{red}\$AT} is {\color{orange}\$OT} means \\{\color{blue}\$AC} is {\color{violet}\$S}\\Literal: \textit{{\color{red}burger} is {\color{orange}loved} means}\\\textit{{\color{blue}food} is {\color{violet}great}}\end{tabular} \\
\hline
\end{tabular}%
}
\caption{The factorized sub-tasks in ABSA. Each of them covers a sub-set of all four prediction targets. \$AT: Aspect Term; \$AC: Aspect Category; \$S: Sentiment; \$OT: Opinion Term; \$TEXT: input text. Both templates and literal values (for \$TEXT = \textit{I loved the burger}) are shown for Output against each task.
}
\label{tab:task_templates}
\end{table*}

\begin{figure}[t]
    \centering
    \includegraphics[height=2.2in]{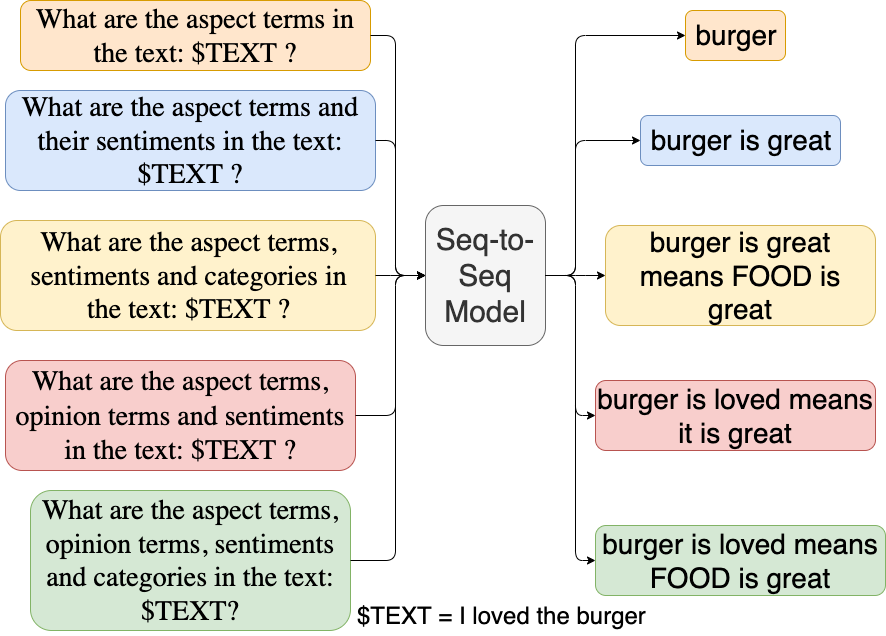}
    \caption{Instruction tuning to solve the sub-tasks related to ABSA. We devise multiple 
    prompts to instruct a seq-to-seq model to learn in multi-task learning manner. 
    }
    \label{fig:main_figure}
\end{figure}

We consider 
multiple factorized sub-tasks involving one or more of the four elements to be predicted. We pose it as a combination of five Question Answering (QA) tasks as illustrated in Figure \ref{fig:main_figure}. For each QA task, an instructional prompt is used to train a seq-to-seq model to learn one or more ABSA elements -- referred to as Instruction Tuning (IT). Our formulation enables learning all sub-tasks via Multi-Task Learning (MTL). 

\ignore{\shuai{General comment: We should give our problem definition (or task formulation) with symbols, where the symbols would be used throughout Section 3, including the subsections below, about input, training, etc.}}

\subsection{Input Transformation}
First, we transform each sentence in the corpus using the instruction templates provided for
each task as shown in Table \ref{tab:task_templates}. Furthermore,
we use multiple paraphrased instruction templates as shown in Table \ref{tab:all_input_prompts}
for a task, and sample randomly when preparing a batch during training (and evaluation) of the seq-to-seq model. However, the target output sequence remains unchanged irrespective of the template sampled for a task.

\begin{table*}[t]
\centering
\resizebox{0.9\textwidth}{!}{%
\begin{tabular}{|l|l|l}
\cline{1-2}
Task                  & Input Prompts                                                                                 &  \\ \cline{1-2}
\multirow{2}{*}{AE}   & Given the text: \$TEXT, what are the aspect terms in it ?                                     &  \\ \cline{2-2}
                      & What are the aspect terms in the text: \$TEXT ?                                               &  \\ \cline{1-2}
\multirow{2}{*}{ASE}  & Given the text: \$TEXT, what are the aspect terms and their sentiments ?                      &  \\ \cline{2-2}
                      & What are the aspect terms and their sentiments in the text: \$TEXT ?                          &  \\ \cline{1-2}
\multirow{4}{*}{TASD} & Given the text: \$TEXT, what are the aspect terms, sentiments and categories ?                &  \\ \cline{2-2}
                      & What are the aspect terms, sentiments and categories in the text: \$TEXT ?                    &  \\ \cline{2-2}
                      & Given the text: \$TEXT, what are the aspect terms, categories and sentiments ?                &  \\ \cline{2-2}
                      & What are the aspect terms, categories and sentiments in the text: \$TEXT ?                    &  \\ \cline{1-2}
\multirow{4}{*}{ASTE} & Given the text: \$TEXT, what are the aspect terms, opinion terms and sentiments ?             &  \\ \cline{2-2}
                      & What are the aspect terms, opinion terms and sentiments in the text: \$TEXT ?                 &  \\ \cline{2-2}
                      & Given the text: \$TEXT, what are the opinion terms, aspect terms and sentiments ?             &  \\ \cline{2-2}
                      & What are the opinion terms, aspect terms and sentiments in the text: \$TEXT ?                 &  \\ \cline{1-2}
\multirow{8}{*}{ASQP} & Given the text: \$TEXT, what are the aspect terms, opinion terms, sentiments and categories ? &  \\ \cline{2-2}
                      & What are the aspect terms, opinion terms, sentiments and categories in the text: \$TEXT ?     &  \\ \cline{2-2}
                      & Given the text: \$TEXT, what are the aspect terms, opinion terms, categories and sentiments ? &  \\ \cline{2-2}
                      & What are the aspect terms, opinion terms, categories and sentiments in the text: \$TEXT ?     &  \\ \cline{2-2}
                      & Given the text: \$TEXT, what are the opinion terms, aspect terms, sentiments and categories ? &  \\ \cline{2-2}
                      & What are the opinion terms, aspect terms, sentiments and categories in the text: \$TEXT ?     &  \\ \cline{2-2}
                      & Given the text: \$TEXT, what are the opinion terms, aspect terms, categories and sentiments ? &  \\ \cline{2-2}
                      & What are the opinion terms, aspect terms, categories and sentiments in the text: \$TEXT ?     &  \\ \cline{1-2}
\end{tabular}
}
\caption{List of input instruction prompts for all the five sub-tasks. \$TEXT is the place holder for actual text.}
\label{tab:all_input_prompts}
\end{table*}

\subsection{Model Training}
Next, we perform 
IT with the seq-to-seq model. We train it in a MTL fashion where input-output combinations are sampled from all tasks simultaneously. We use the following loss for model training:
\vspace{-0.5cm}
\begin{equation}
\mathcal{L} = -\frac{1}{T}\sum_{t=1}^{T}\sum_{i=1}^{n} \log p_{\theta}(y_i|y_1,..., y_{i-1}, \mathbf{x_t}).
\end{equation}
 where $\mathbf{x_t}$
 is the transformed input sequence ($\mathbf{x}$) for $t^{\text{th}}$ task. $\theta$ is the set of model parameters. $n$ is the length of output sequence. $y_i$ is the $i^{\text{th}}$ token in output sequence. $T$ is the number of tasks. The model parameters are updated using {\tt Adam} optimizer with weight decay \cite{Loshchilov2019DecoupledWD}.

\subsection{Output Transformation}
Finally, we transform the output using the templates provided in the rightmost column in Table \ref{tab:task_templates}. 
In case there is more than one quadruple in the output, we use a special separation token {\tt[SSEP]}. We map sentiment classes positive, negative and neutral to \textit{great, bad} and \textit{ok} respectively in the output similar to \cite{zhang-etal-2021-aspect-sentiment}.
During inference, we apply the reverse transformations to recover the quadruples for evaluation.

\ignore{\shuai{This paragraph seems a bit disconnected and mixed of things and not clear. (We can improve later if we don't have time now, but we need to improve it. BTW- we still haven't talk about our MLT method.)}}



 \ignore{\shuai{We need to have a subsection to talk about multi-task learning (MTL), i.e., our final model (using MTL). Also how it works.}}


%% file: experiments.tex
\section{Experiments}
\label{sec:experiments}
As this work is one of the first few attempts
towards studying few-shot learning in ABSA context, 
unsurprisingly, there is a lack of standard few-shot datasets. We emulate few-shot data drawing inspiration from the literature \cite{halder2020coling, ma2022label} for our experiments.

\subsection{Datasets: Few-shot Preparation}
We use three datasets, \textsc{Rest15}, \textsc{Rest16} from \cite{zhang-etal-2021-aspect-sentiment} and \textsc{Laptop14} from \cite{xu-etal-2020-position}.
For the first two, we shuffle the data with fixed random seed, and select first few samples so that
there are at least $k$ samples from each aspect category\footnote{It is not feasible to guarantee exactly $k$ samples since an example can have multiple aspect categories. \cite{ma2022label}}.
As \textsc{Laptop14} does not have aspect category annotations, we select $k$ examples per sentiment class instead, following the same principle (statistics in Table \ref{tab:dataset_stats}).

\subsection{Baselines and Models for Comparison}
As a strong baseline, we consider \textsc{Paraphrase} (or \textsc{Para}) model\footnote{Other competitive models can be found in \cite{zhang-etal-2021-aspect-sentiment}. Since \textsc{Para} has outperformed them, we focus on it.}
-- the current state-of-the-art for TASD, ASTE, and ASQP tasks \cite{zhang-etal-2021-aspect-sentiment}. It uses the same backbone model as of ours, which ensures fair comparison. However, for the other two tasks \textsc{Para} is not applicable, hence we use a generative framework called \textsc{BartAbsa} as the baseline \cite{yan-etal-2021-unified}. All the \textsc{Para} numbers are obtained using our implementation for a fair comparison (\textit{cf} Section \ref{appendix:implementation}).

To understand the impact of all the components in our approach, we consider two model ablations:
\begin{enumerate}
\setlength{\itemsep}{0em}
\item \textbf{Text}: \$TEXT is directly used as input 

\item \textbf{IT}: \$TEXT is transformed to instructions
\end{enumerate}



We refer to our full proposed model as \textbf{IT-MTL}, it covers all the tasks. Table \ref{tab:ablation_illustration} provides illustrations of the input prompts for the ablations.

\begin{table}[]
\centering
\resizebox{0.48\textwidth}{!}{%
\begin{tabular}{|c|c|}
\hline
\textbf{Ablation} & \textbf{Input Prompt} \\ \hline
Text & \$TEXT \\ \hline
IT & What are the aspect terms in the text: \$TEXT? \\ \hline
\multirow{5}{*}{IT-MTL} & What are the aspect terms in the text: \$TEXT? \\ \cline{2-2} 
 & \begin{tabular}[c]{@{}c@{}}What are the aspect terms and their sentiments \\ in the text: \$TEXT?\end{tabular} \\ \cline{2-2} 
 & \begin{tabular}[c]{@{}c@{}}Given the text: \$TEXT, what are the aspect \\ terms, sentiments and categories?\end{tabular} \\ \cline{2-2} 
 & \begin{tabular}[c]{@{}c@{}}Given the text: \$TEXT, what are the aspect \\ terms, opinion terms and sentiments?\end{tabular} \\ \cline{2-2} 
 & \begin{tabular}[c]{@{}c@{}}What are the aspect terms, opinion terms, \\ sentiments and categories in the text: \$TEXT ?\end{tabular} \\ \hline
\end{tabular}%
}
\caption{Illustration of input prompts to the seq-to-seq model for various ablations of our proposed approach.}
\label{tab:ablation_illustration}
\end{table}



\subsection{Experimental Setup}
We use {\tt t5-base}
\cite{JMLR:v21:20-074} as the backbone for our models. Results are averaged over $5$ runs with random seeds (\textit{cf} Section \ref{appendix:hyperparams} for all details). Micro F1 is the evaluation metric following previous work \cite{zhang-etal-2021-aspect-sentiment}.

\subsection{Results}
\ignore{\shuai{We can also consider showing full shot results first here , Table \ref{tab:all_task_results}. And then Few Shot. Table 2 and 3}
\robert{Overall, adding instructions gives $1.7\%$ improvement across all settings and all datasets when compared to the text-only counterpart. Adding MTL further improves the results, bringing the overall improvement to $2.58\%$.}}

We present results for all the datasets in Table \ref{tab:all_task_results}. Since, \textsc{Laptop14} lacks aspect category annotations, TASD and ASQP are not applicable. We make four key observations from the results.\\


\noindent{\textbf{Ablation Study:}} First, IT beats Text in most settings proving effectiveness of our instructions. 
Second, we observe that IT-MTL outperforms others on \textsc{Rest15}, and \textsc{Rest16} substantially in few-shot settings, except on \textsc{Laptop14} as IT-MTL underperforms on AE task. This might be attributed to the absence of TASD, ASQP tasks. Overall, we observe the trend IT-MTL $>$ IT $>$ Text.


\begin{table}[]
    \begin{subtable}[c]{.43\textwidth}
        \centering
        \resizebox{1\textwidth}{!}{%
            \begin{tabular}{|c|c|c|c|c|c|}
            \hline
             \textbf{Task} &  \textbf{Model}   & \textbf{K=5}  & \textbf{K=10}  & \textbf{K=20} & \textbf{K=50}
            \\
            \hline
            \multirow{4 }{*}{AE}  & \textsc{BartAbsa} & 19.68 &	42.99 &	57.43 &	63.48\\
            \cline{2-6}
            & Text & 43.95 & 54.38 & 59.75 & 61.75\\
            & IT & \textbf{45.24} & \underline{55.1} & \underline{60.33} & \underline{64.15}\\
            & IT-MTL & \underline{44.18} & \textbf{56.57} & \textbf{62.65} & \textbf{67.22}\\
            \hline
            \multirow{4 }{*}{AESC} & \textsc{BartAbsa} & 10.77 &	27.38 &	42.23 &	52.55\\
            \cline{2-6}
            & Text & 37.33 & \underline{47.68} & 50.6 & 56.69\\
            & IT & \textbf{39.4} & \textbf{49.43} & \underline{52.06} & \underline{58.4}\\
            & IT-MTL & \underline{38.99} & 47.62 & \textbf{53.58} & \textbf{59.54}\\
            \hline
            \multirow{4 }{*}{TASD} & \textsc{Para.} & 21.34 & \textbf{37.39} & 42.52 & 47.57\\
            \cline{2-6} 
            & Text & 22.55 & 36.37 & 42.28 & 48.52\\
            & IT & \underline{22.92} & 36.52 & \underline{43.2} & \underline{50.14}\\
            & IT-MTL & \textbf{27.05} & \underline{36.81} & \textbf{43.56} & \textbf{50.24}\\ \hline
            \multirow{4 }{*}{ASTE} & \textsc{Para.} & 22.07 & \underline{32.49} & 36.28 & 41.12 \\
            \cline{2-6} 
            & Text & 18.49 & 30.17 & 35.66 & 41.49\\
            & IT & \underline{22.38} & 32.11 & \underline{36.67} & \underline{41.65}\\
            & IT-MTL & \textbf{22.7} & \textbf{33.52} & \textbf{37.78} & \textbf{43.84}\\
            \hline
            \multirow{4 }{*}{ASQP} & \textsc{Para.} & \underline{13.65} & 22.90 & 27.87 & 34.49\\
            \cline{2-6} 
            & Text & 12.15 & 22.19 & 28.82 & 33.96\\
            & IT & 13.3 & \underline{24.35} & \underline{29.66} & \underline{36.78}\\
            & IT-MTL & \textbf{15.54} & \textbf{25.46} & \textbf{31.47} & \textbf{37.72}\\
            \hline
            \end{tabular}
        }
        \caption{\textsc{Rest15}}
        \label{tab:results_rest15}
    \end{subtable}
    \begin{subtable}[c]{.43\textwidth}
    \centering
    \resizebox{1\textwidth}{!}{%
        \begin{tabular}{|c|c|c|c|c|c|}
            \hline
             \textbf{Task} &  \textbf{Model}   & \textbf{K=5}  & \textbf{K=10}  & \textbf{K=20} & \textbf{K=50} 
            \\
            \hline
            \multirow{4}{*}{AE} & \textsc{BartAbsa} & 31.48 & 55.90 & 62.96 & \underline{71.06}\\
            \cline{2-6}  & Text & 52.7 & 58.5 & 61.49 & 67.21\\
            & IT & \underline{55.64} & \underline{59.36} & \underline{63.75} & 68.14\\
            & IT-MTL & \textbf{59.41} & \textbf{61.87} & \textbf{66.88} & \textbf{71.18}\\
            \hline
            \multirow{4 }{*}{AESC} & \textsc{BartAbsa} & 25.45 & 46.31 & 53.27 & 62.90\\
            \cline{2-6}
            & Text & 49.13 & 54.54 & 57.05 & 62.75\\
            & IT & \underline{51.93} & \underline{55.29} & \underline{59.96} & \underline{63.45}\\
            & IT-MTL & \textbf{52.42} & \textbf{55.37} & \textbf{60.22} & \textbf{65.14}\\
            \hline
            \multirow{4 }{*}{TASD} & \textsc{Para.} & 28.93 & \underline{38.99} & \underline{48.29} & 54.89\\
            \cline{2-6} 
            & Text & 30.65 & 38.39 & 46.72 & 54.04\\
            & IT & \underline{34.38} & 38.58 & 47.66 & \underline{55.16}\\
            & IT-MTL & \textbf{40.45} & \textbf{42.41} & \textbf{48.83} & \textbf{55.82}\\ \hline
            \multirow{4 }{*}{ASTE} & \textsc{Para.} & 32.48 & 38.90 & 43.51 & 51.47\\
            \cline{2-6} 
            & Text & 28.44 & 38.23 & 42.12 & 50.9\\
            & IT & \underline{33.08} & \textbf{41.12} & \underline{44.08} & \underline{51.69}\\
            & IT-MTL & \textbf{35.75} & \underline{38.95} & \textbf{44.75} & \textbf{52.94}\\
            \hline
            \multirow{4 }{*}{ASQP} & \textsc{Para.} & 20.02 & 28.58 & 36.26 & 43.50\\
            \cline{2-6} 
            & Text & 20.98 & 28.06 & 35.04 & 45.26\\
            & IT & \underline{23.86} & \underline{30.02} & \underline{37.20} & \underline{46.9}\\
            & IT-MTL & \textbf{27.02} & \textbf{31.66} & \textbf{38.06} & \textbf{47.48}\\
            \hline
        \end{tabular}
        }
        \caption{\textsc{Rest16}}
        \label{tab:results_rest16}
    \end{subtable}
    \begin{subtable}[c]{.43\textwidth}
    \centering
    \resizebox{1\textwidth}{!}{%
        \begin{tabular}{|c|c|c|c|c|c|}
            \hline
             \textbf{Task} &  \textbf{Model}   & \textbf{K=5}  & \textbf{K=10}  & \textbf{K=20} & \textbf{K=50} 
            \\
            \hline
            \multirow{4}{*}{AE} & \textsc{BartAbsa} & -- & 5.54 & 33.04 & 60.98 \\
            \cline{2-6}
            & Text & \textbf{34.64} & 42.26 & 51.11 & 59.62\\
            & IT & \underline{34.29} & \textbf{47.4} & \underline{52.39} & \textbf{63.86}\\
            & IT-MTL & 31.54 & \underline{42.73} & \textbf{53.08} & \underline{63.71}\\ \hline
            \multirow{4}{*}{AESC} & \textsc{BartAbsa} & -- & 4.75 & 24.92 & 50.01\\
            \cline{2-6}
            & Text & 21.68 & 30.7 & 37.74 & 50.39\\
            & IT & \underline{23.28} & \textbf{36.55} & \underline{43.39} & \underline{52.92}\\
            & IT-MTL & \textbf{25.01} & \underline{34.44} & \textbf{44.5} & \textbf{53.75}\\ \hline
            \multirow{4 }{*}{ASTE} & \textsc{Para.} & \textbf{14.99} & \underline{23.87} & \underline{30.12} & \textbf{43.75}\\
            \cline{2-6} 
            & Text & 10.10 & 16.27 & 26.37 & 39.65\\
            & IT & 12.60 & 21.31 & 30.03 & 41.91\\
            & IT-MTL & \underline{14.18} & \textbf{24.09} & \textbf{32.39} & \underline{42.62}\\
            \hline
        \end{tabular}
        }
        \caption{\textsc{Laptop14}}
        \label{tab:results_laptop14}
    \end{subtable}
\caption{Comparison of IT-MTL with baselines. \textbf{Bolded:} best, \underline{Underlined:} second-best. 
`--' denotes the model failed to obtain a non-zero score.}

\label{tab:all_task_results}
\end{table}



\noindent{\textbf{Baseline Comparison:}} 
Third, our proposed IT-MTL approach outperforms \textsc{Para}, and \textsc{BartAbsa}
comfortably in most few-shot settings across all datasets with a performance boost of $8.29$ F1 on average. We observe some exceptions in \textsc{Laptop14}, where \textsc{Para} outperforms IT-MTL slightly on ASTE -- possibly due to the missing tasks that involve aspect category annotations. Fourth, we also experiment with the full training datasets and summarize them in Figure \ref{fig:full_comparison}. In $4$ out of $5$ tasks, our IT-MTL model either outperforms or does at par with the SOTA baselines. Interestingly, in case of AE, it falls behind \textsc{BartAbsa} by ~3.5 F1 scores. We attribute this difference to the advanced decoding strategies used in \textsc{BartAbsa} which are orthogonal to our work.



Regarding the randomness introduced by the seeds, we observe that the model training
is reasonably stable across tasks (\textit{cf} Table \ref{tab:results_with_std}).
Overall, we conclude that in few-shot settings, our proposed IT-MTL leverages the knowledge from multiple tasks, and improves the generalization of the underlying seq-to-seq model across all the ABSA tasks.

%% file: conclusion.tex
\section{Conclusion}
\label{sec:conclusion}

In this paper, we posed ABSA as an instruction tuning based seq-to-seq modeling task. We factorized the overall quadruple prediction task into five sub-tasks resembling Question Answering tasks. We proposed a multi-task learning based approach using a pre-trained seq-to-seq model.
\ignore{\robert{Add that the instruction tuning is something proposed here as well}} 
We experimented with customer reviews from two domains, showed that our approach gives superior performance compared to baseline models in few-shot, and stays comparable in full fine-tuning scenarios. 

%% file: limitations.tex
\section{Limitations}
\label{sec:limitations}

First, our work essentially relies upon a generative language model to understand the relationships between the sentiment elements in contrast to discriminative/extractive models which make structured predictions by design. As a result, our model is susceptible to usual anomalies suffered by generative models \textit{e.g.,} malformed outputs. We recover the quadruples from the model's output sequence using regular expression based matching with fixed templates, as a result, an end-user will never receive any irrelevant text generated by the model. However, the accuracy will still be impacted in such cases nevertheless. Second, input sequences in user-generated content can be arbitrarily long and that might result in increased decoding time because of the underlying generative model. Last but not the least, all the instruction templates we provide in this work are designed solely for English. It would be interesting to explore systematic ways to be more language inclusive for instruction tuning based ABSA.

%% file: appendix.tex
\section{Appendix}

\subsection{List of input instruction prompts}

\subsection{Hyperparameters}
\label{appendix:hyperparams}

We set the learning rate to 3e-4 for all the experiments in this paper. We train each model for a fixed number of 20 epochs similar to \citeauthor{zhang-etal-2021-aspect-sentiment}. For full-shot experiments, we use a batch size of 16. For $k$=5, 10, 20 and 50 we use a batch size of 2, 2, 4 and 8 respectively. The maximum sequence length is set to 160. Longer sequences are truncated and shorter sequences are padded. Finally, we use Adam optimizer with weight decay.

\subsection{Dataset Statistics}
\label{appendix:dataset_stats}

Table \ref{tab:dataset_stats} presents the number of sentences in each dataset. Please note that for \textsc {Laptop14} dataset, the few-shot data for different values of K was selected based on sentiment classes instead of Aspect category due to lack of category annotations.

\begin{table*}[]
\centering
\resizebox{1.0\textwidth}{!}{%
\begin{tabular}{|c|ccccc|ccccc|ccccc|}
\hline
 & \multicolumn{5}{c|}{Rest15} & \multicolumn{5}{c|}{Rest16} & \multicolumn{5}{c|}{Laptop14} \\ \hline
 & \multicolumn{1}{c|}{K=5} & \multicolumn{1}{c|}{K=10} & \multicolumn{1}{c|}{K=20} & \multicolumn{1}{c|}{K=50} & Full & \multicolumn{1}{c|}{K=5} & \multicolumn{1}{c|}{K=10} & \multicolumn{1}{c|}{K=20} & \multicolumn{1}{c|}{K=50} & Full & \multicolumn{1}{c|}{K=5} & \multicolumn{1}{c|}{K=10} & \multicolumn{1}{c|}{K=20} & \multicolumn{1}{c|}{K=50} & Full \\ \hline
Train & \multicolumn{1}{c|}{25} & \multicolumn{1}{c|}{46} & \multicolumn{1}{c|}{86} & \multicolumn{1}{c|}{181} & 834 & \multicolumn{1}{c|}{22} & \multicolumn{1}{c|}{43} & \multicolumn{1}{c|}{77} & \multicolumn{1}{c|}{179} & 1264 & \multicolumn{1}{c|}{11} & \multicolumn{1}{c|}{19} & \multicolumn{1}{c|}{40} & \multicolumn{1}{c|}{106} & 906 \\ \hline
Dev & \multicolumn{1}{c|}{21} & \multicolumn{1}{c|}{35} & \multicolumn{1}{c|}{68} & \multicolumn{1}{c|}{140} & 209 & \multicolumn{1}{c|}{26} & \multicolumn{1}{c|}{42} & \multicolumn{1}{c|}{73} & \multicolumn{1}{c|}{159} & 316 & \multicolumn{1}{c|}{8} & \multicolumn{1}{c|}{16} & \multicolumn{1}{c|}{34} & \multicolumn{1}{c|}{86} & 219 \\ \hline
Test & \multicolumn{5}{c|}{537} & \multicolumn{5}{c|}{544} & \multicolumn{5}{c|}{328} \\ \hline
\end{tabular}
}
\caption{Number of sentences in each dataset. The same test set was used for few-shot and full-shot evaluation.}
\label{tab:dataset_stats}
\end{table*}

\subsection{Results on Full Datasets}
\begin{figure}[]
    \centering
    \includegraphics[height=1.5in]{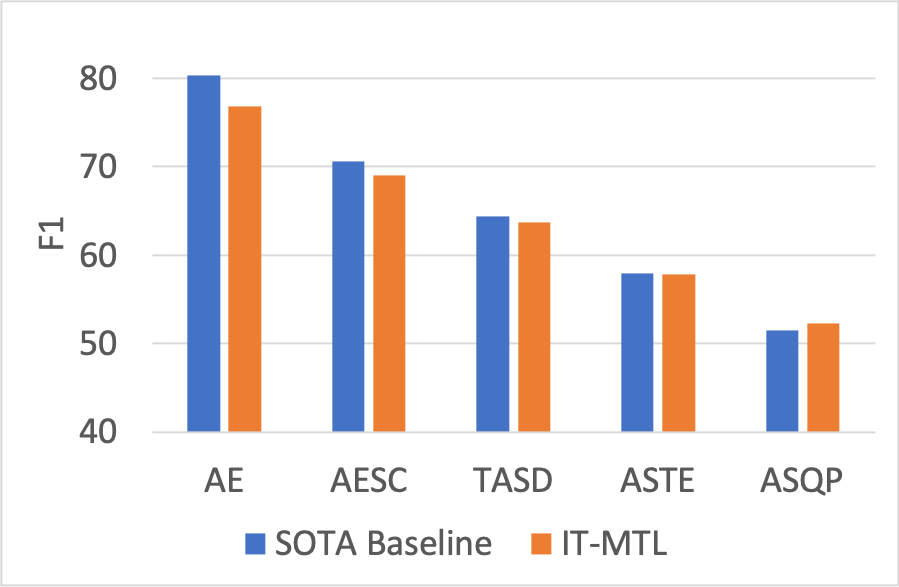}
    \caption{The average F1 scores achieved by our IT-MTL model and the relevant baseline. For AE, AESC the baseline is BARTABSA, and for others \textsc{Para.} is the baseline.}
    \label{fig:full_comparison}
\end{figure}

The averaged results across full datasets (\textsc{Rest15}, \textsc{Rest16} and \textsc{Laptop14} ) are in Figure \ref{fig:full_comparison}.

\subsection{Implementation Issues}
\label{appendix:implementation}
We extend \newcite{zhang-etal-2021-aspect-sentiment}'s library to implement our models. A careful reader might notice that the \textsc{Para} and our text-only ablation should be similar as the only difference is in the output prompts. However, in practice we observe a large gap in few-shot performance between these two when we obtain the numbers for \textsc{Para} with authors' published sources. Upon investigating, we discovered a few implementation issues in their sources. Our implementation improves \textsc{Para}'s F1 scores in few-shot settings and we report that to ensure a fair comparison. It brings the gap down from $6.75$ to $2.32$ in terms of absolute F1 scores between IT-MTL and \textsc{Para}.

\noindent{\textbf{Evaluation Logic:} We observe another critical issue in the evaluation logic in \citeauthor{zhang-etal-2021-aspect-sentiment}'s sources\footnote{\url{https://github.com/IsakZhang/ABSA-QUAD/blob/master/eval_utils.py\#L90}}. It discounts the repetitions of the same tuple produced in the output. For illustration, let us assume for a review the target tuples for AE task are \textit{burger, fries}. Now, if the seq-to-seq model outputs \textit{burger, burger}, the logic in their sources computes the true positive count to be $2$, whereas it should be only $1$. This ultimately leads to an inflated F1 score. We fix this issue in our evaluation and comparisons with \textsc{Para}. The reported F1 for \textsc{Para} with the original logic was $61.13$, after the fix it becomes $60.70$ on full corpus of \textsc{Laptop14}. Overall, we observe that for few-shot cases, this issue becomes more apparent compared to the high-shot ones.}


\begin{table*}[]
\centering
\resizebox{0.65\textwidth}{!}{%
    \begin{tabular}{|l|l|c|c|c|c|}
    \hline
    \multicolumn{1}{|c|}{Dataset} & Model  & K=5                              & K=10                             & K=20                             & K=50                             \\ \hline
    \multirow{2}{*}{\textsc{Rest15}}       & \textsc{Para.}   & \multicolumn{1}{l|}{13.65$\pm$0.92} & \multicolumn{1}{l|}{22.90$\pm$0.50} & \multicolumn{1}{l|}{27.87$\pm$1.64} & \multicolumn{1}{l|}{34.49$\pm$0.64} \\ \cline{2-6} 
                                  & IT-MTL & 15.54$\pm$1.61                     & 25.46$\pm$1.09                     & 31.47$\pm$0.58                      & 37.72$\pm$0.76                      \\ \hline
    \multirow{2}{*}{\textsc{Rest16}}       & \textsc{Para.}   & \multicolumn{1}{l|}{20.02$\pm$1.43} & \multicolumn{1}{l|}{28.58$\pm$1.41} & \multicolumn{1}{l|}{36.26$\pm$0.54} & \multicolumn{1}{l|}{43.50$\pm$0.29} \\ \cline{2-6} 
                                  & IT-MTL & 27.02$\pm$1.29                      & 31.66$\pm$1.39                      & 38.06$\pm$1.69                      & 47.48$\pm$1.20                      \\ \hline
    \end{tabular}
}
\caption{Results (F1 $\pm$ standard deviation) for ASQP task. The F1 scores remain reasonably stable with the standard deviation being under $\sim$1.6 F1 points in all cases.}
\label{tab:results_with_std}
\end{table*}

\subsection{Stochasticity in Few-shot Data Sampling}
So far, we keep the few-shot data fixed and vary the seed $5$ times. To observe the effect of another form of stochasticity, in Table \ref{tab:results_with_diff_samples}, we sample few-shot data $5$ times for \textsc{Rest16} and keep the seed fixed. We observe that the trend remains the same.

\begin{table}[H]
\centering
\resizebox{0.33\textwidth}{!}{%
    \begin{tabular}{|c|c|c|c|c|}
    \hline
    \textbf{Model}                                             & \textbf{K=5}            & \textbf{K=10}           & \textbf{K=20}           & \textbf{K=50}           \\ \hline
    Text                                              & 21.99          & 29.3           & 37.92          & 46.83          \\ 
    IT                                                & 22.91          & 31.24          & 38.00          & 47.94          \\ 
    IT-MTL                                                & \textbf{24.97}          & \textbf{32.25}          & \textbf{39.89}          & \textbf{48.20}          \\ \hline
    \end{tabular}
}
\caption{ASQP Results for \textsc{Rest16} averaged across $5$ different k-shot samples.}
\label{tab:results_with_diff_samples}
\end{table}